\documentclass[runningheads]{llncs}

\usepackage[T1]{fontenc}

\usepackage{url}
\usepackage{graphicx}
\usepackage{amsmath}
\usepackage{amssymb}
\usepackage[colorlinks, linkcolor=blue, anchorcolor=blue, citecolor=blue]{hyperref}
\usepackage{adjustbox}

\usepackage{subcaption}
\usepackage{array}
\newcolumntype{L}[1]{>{\raggedright\let\newline\\\arraybackslash\hspace{0pt}}m{#1}}
\newcolumntype{C}[1]{>{\centering\let\newline\\\arraybackslash\hspace{0pt}}m{#1}}
\newcolumntype{R}[1]{>{\raggedleft\let\newline\\\arraybackslash\hspace{0pt}}m{#1}}

\usepackage{graphicx,verbatim}

\usepackage{booktabs}
\usepackage[capitalize]{cleveref}
\usepackage{multirow}

\usepackage[table]{xcolor}
\definecolor{lightgray}{gray}{0.9}

\newcommand{\bfx}{\mathbf{x}}
\newcommand{\bfc}{\mathbf{c}}
\newcommand{\bfr}{\mathbf{r}}
\newcommand{\bfy}{\mathbf{y}}
\newcommand{\bfn}{\mathbf{n}}
\newcommand{\bfa}{\mathbf{a}}
\newcommand{\bfb}{\mathbf{b}}

\newcommand{\calN}{\mathcal{N}}
\newcommand{\calA}{\mathcal{A}}
\newcommand{\calB}{\mathcal{B}}
\newcommand{\calL}{\mathcal{L}}
\newcommand{\bbR}{\mathbb{R}}

\makeatletter
\def\thanks#1{\protected@xdef\@thanks{\@thanks
        \protect\footnotetext{#1}}}
\def\@makefntext#1{
  \noindent
  \hb@xt@\z@{\hss\@makefnmark}#1}
\makeatother

\begin{document}
\title{Learning Concept-Driven Logical Rules\texorpdfstring{\\}{ } for Interpretable and Generalizable Medical Image Classification}
\titlerunning{Learning Concept-Driven Logical Rules}

\author{Yibo Gao$^{1}$ \and Hangqi Zhou$^{1}$ \and Zheyao Gao$^{2}$ \and Bomin Wang$^{1}$ \and \\ Shangqi Gao$^{3}$ \and Sihan Wang$^{1}$ \and Xiahai Zhuang$^{1}$}
\authorrunning{Y. Gao et al.}
\institute{School of Data Science, Fudan University \and Dept. of Computer Science and Engineering, The Chinese University of Hong Kong \and Dept. of Oncology, University of Cambridge}

\maketitle

\vspace{-3mm}

\begin{abstract}
The pursuit of decision safety in clinical applications highlights the potential of transparent methods in medical imaging. While concept-based models offer local concept explanations (instance-level), they often neglect the global decision logic (dataset-level). Moreover, these models often suffer from \textit{concept leakage}, where unintended information within soft concept representations undermines both interpretability and generalizability. To address these limitations, we propose \textbf{Concept Rule Learner} (CRL), a novel framework to learn Boolean logical rules from binary visual concepts. CRL employs logical layers to capture concept correlations and extract clinically meaningful rules, thereby providing both local and global interpretability. 
The results from two tasks demonstrate that CRL achieves competitive performance with existing interpretable methods while improving generalizability to out-of-distribution data. The code of our work is available at \url{https://github.com/obiyoag/crl}.

\keywords{Explainable-AI \and Concept Learning \and Rule-based Model.}

\end{abstract}

\vspace{-5mm}
\section{Introduction}
Deep learning models, especially those operating as black boxes, have shown great promise in medical imaging applications~\cite{nnUNet,med_sam,med_llava}. However, the high standards of trust and accountability required in healthcare have spurred growing interest in transparent models~\cite{medical_XAI,stop_explain}, where "\textit{explainability}" and "\textit{logic}" have been emphasized as two aspects by the FDA principles~\cite{us2024transparency}. "\textit{Logic}" refers to the decision rules underlying model predictions, similar to clinical practice, where medical professionals assess symptoms and make decisions based on established clinical guidelines or rules. While recent research increasing focuses on "\textit{explainability}" via concept explanations, less attention has been paid to "\textit{logic}" rules for medical imaging applications.

Concept Bottleneck Models (CBMs)~\cite{cbm} are a prevalent framework for providing concept explanations. In CBMs, a concept predictor generates explanable concepts, which are then used by a label predictor to make final predictions. Based on CBMs, Concept Embedding Models (CEMs)~\cite{cem} enhance predictive performance by employing high-dimensional concept embeddings.
However, these models only provide explicit local explanations by focusing on individual predictions, detailing how predicted concepts influence each decision~\cite{lf_cbm,labo}. \textit{They offer insights for particular instances but may not fully capture the overall decision logic across the entire medical dataset.} For more comprehensive interpretability, integrating both local concept explanations and global logical rules is essential~\cite{interpretable,local_global} for transparent medical decision-making.
A recent work, named Deep Concept Reasoner (DCR)~\cite{dcr}, explores to extract syntactic logical rules from concept embeddings. However, because DCR relies on fuzzy logic and concept embeddings, its decision-making process is less transparent.

Moreover, the above concept-based models suffer from \textit{concept leakage}~\cite{leakage}, where the label predictors inadvertently exploit unintended image information from soft concepts (\textit{i.e.}, probabilities or embeddings). \textit{The leakage compromises both interpretability and generalizability}~\cite{leakage2}. For interpretability, predictions are influenced not only by the intended concepts but also by image information encoded within the soft concept representations. The concept predictor no longer needs to faithfully predict the concepts for accurate label predictions. Regarding generalizability, reliance on leaked information may lead to overfitting, rendering the models less robust to distribution shifts. Hard CBMs, which use binary concepts (\textit{i.e.}, 0 and 1) for the label predictor, reduce leakage as these binary values inherently carry less extraneous information. However, they often exhibit poor performance as they predict concepts independently, neglecting the correlations between concepts.

To address the above limitations and achieve transparent medical decision-making, we propose \textbf{Concept Rule Learner} (CRL), a framework to learn Boolean logical rules from medical data. Inspired by prior works on neuro-symbolic learning~\cite{deeplogic,rrl,nlil}, CRL extends logical layers to capture correlations between binary visual concepts, mitigating the issue of concept leakage. Each logical layer comprises a conjunction layer and a disjunction layer, which perform \texttt{AND} and \texttt{OR} operations, respectively. The connections across these layers generate flexible decision rules, and the final prediction for an input image is derived by aggregating the contributions of triggered rules through linear weights. By making decisions based on domain-invariant logical rules, CRL not only incorporates global interpretability, but also improves generalizability to unseen domains. Our main contributions can be summarized as follows:
\begin{itemize}
    \item[$\bullet$] We propose CRL, a novel framework that learns Boolean logical rules from binary visual concepts to model concept correlations, while mitigating the issue of concept leakage.
    \item[$\bullet$] CRL not only delivers concept explanations for individual predictions but also extracts decision rules for the entire datasets, thereby unifying local and global interpretations.
    \item[$\bullet$] We evaluate the effectiveness of the proposed method on two medical image classification tasks. The experimental results demonstrate that our approach could extract meaningful concept logical rules and exhibits superior generalizability to the unseen dataset.
\end{itemize}

\section{Method}

\subsection{Model Architecture}

\begin{figure}[t]
  \centering
   \includegraphics[width=\linewidth]{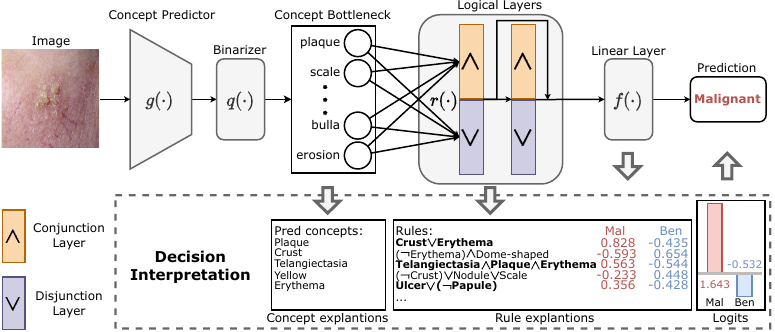}
   \caption{The architecture of Concept Rule Learner (CRL). CRL is a composition of four functions $(f\circ r \circ q\circ g)$, where $g(\cdot)$ represents the concept predictor, $r(\cdot)$ denotes a stack of logical layers, $q(\cdot)$ means discretization and $f(\cdot)$ corresponds to a linear layer. The decision interpretation is presented in the dashed box, illustrating how concepts and rules contribute to the final prediction.}
   \label{fig:method_overview}
\end{figure}

Given an image $\bfx\in\bbR^{n}$ annotated with a task label $y\in\{1, \cdots, L\}$ and $K$ concept labels $\bfc\in\{0, 1\}^{K}$, our objective is to develop a \textbf{Concept Rule Learner} (CRL) capable of performing medical image classification through concept-driven logical rules. As illustrated in \cref{fig:method_overview}, the CRL framework is \textbf{a quadruple of functions} $\boldsymbol{(g, q, r, f)}$, whose composition $(f\circ r \circ q\circ g)$ predicts the label based on the derived concepts and rules.
\textbf{The first function} $\boldsymbol{g}:\bbR^{n}\to[0,1]^{K}$, referred to as the concept predictor, learns a mapping from the input image $\bfx$ to a set of concept activations $\hat{\bfc}=g(\bfx)\in[0, 1]^{K}$, where $\hat{c}_i$ represents the probability supporting the presence of the $i$-th concept.
\textbf{The second function} $\boldsymbol{q}: [0,1]^{K}\to\{0,1\}^{K}$ serves as a binarizer that converts the concept activations into binary values.
\textbf{The third function} $\boldsymbol{r}:\{0, 1\}^{K}\to\{0, 1\}^{R}$, consists of a series of logical layers, mapping the activated concepts to a set of rule activations $\bfr=r(q(\hat{\bfc}))\in\{0,1\}^{R}$, where $R$ denotes the total number of extracted logical rules. A rule activation $r_i=1$ indicates that the $i$-th rule is activated for the input image $\bfx$ and $r_i=0$ indicates otherwise.
\textbf{The fourth function} $\boldsymbol{f}:\{0, 1\}^{R}\to\bbR^{L}$, represents the final linear layer, which learns the mapping from the activated logical rules to the task logits $\hat{\bfy}=f(\hat{\bfr})\in\bbR^{L}$.

\begin{figure}[t]
    \centering
    \begin{subfigure}[b]{0.55\textwidth}
      \includegraphics[width=\linewidth]{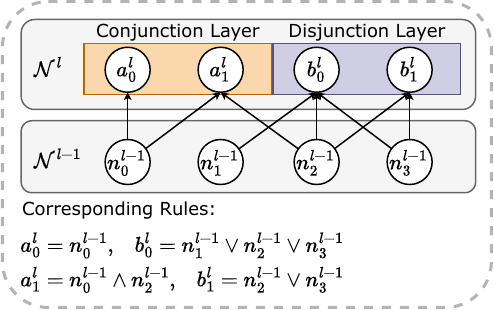}
      \caption{}
      \label{fig:logical_layer_example}
    \end{subfigure}
    \hfill
    \begin{subfigure}[b]{0.40\textwidth}
      \includegraphics[width=\linewidth]{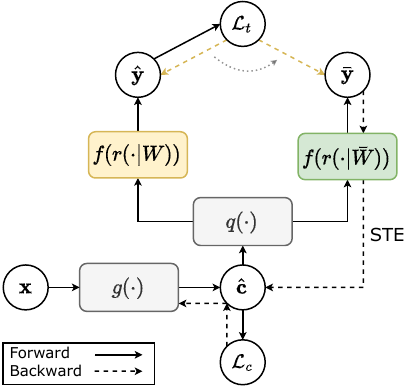}
      \caption{}
      \label{fig:compute_graph}
    \end{subfigure}
    \caption{(a) An example of two adjacent logical layers, where the directed arrows indicates the presence of the connections between nodes. The corresponding rules could be derived by analyzing the connections. (b) The computational graph of CRL. Arrows with solid lines represent forward pass while arrows with dashed lines represent backpropagation.}
\end{figure}

\subsection{Logical Operation Modelling}
To learn logical rules from concept activations, the function $r(\cdot)$ adopts a series of logical layers to model the concept-based rules with Boolean logic. Let $\calN^{l}$ denote the $l$-th logical layer, where $n_j^{l}$ represents the $j$-th node within the layer. The output of layer $\calN^{l}$ is a vector containing the values of all nodes, denoted as $\bfn^{l}$. As shown in \cref{fig:logical_layer_example}, each logical layer $\calN^{l}$ consists of a conjunction layer and a disjunction layer, denoted by $\calA^{l}$ and $\calB^{l}$, respectively. The $i$-th node in the conjunction layer, denoted as $a_i^{l}$, represents the conjunction (\texttt{AND} operation) of nodes from the preceding layer connected to it. Conversely, the $i$-th node in the disjunction layer, denoted as $b_i^{l}$, encapsulates the disjunction (\texttt{OR} operation) of its connected predecessors. Formally, the nodes in the conjunction and disjunction layers are defined as:
\begin{equation}
\label{eq:discrete_logic}
a_{i}^{(l)} =
\bigwedge_{W_{i,j}^{(l,0)}=1}n_{j}^{l-1},\;\;\;\;\;\; b_{i}^{l} = \bigvee_{W_{i,j}^{(l,1)}=1}n_{j}^{l-1},
\end{equation}
where $W^{(l,0)}$ denotes the adjacency matrix of the conjunction layer $\mathcal{A}^{l}$ and the previous layer $\mathcal{N}^{l-1}$, with $W_{i,j}^{(l,0)} \in \{0,1\}$. Specifically, $W_{i,j}^{(l,0)}=1$ indicates the presence of an edge connecting $a_{i}^{l}$ to $n_{j}^{l-1}$, while $W_{i,j}^{(l,0)}=0$ indicates the absence. Similarly, $W^{(l,1)}$ is the adjacency matrix of the disjunction layer $\mathcal{B}^{l}$ and $\mathcal{N}^{l-1}$. These adjacency matrices are treated as the weight matrices for the logical layers, analogous to weight matrices in neural networks. The output of the $l$-th layer is the concatenation of the outputs of the conjunction and disjunction layers, \textit{i.e.}, $\bfn^{l}=\bfa^{l}\oplus \bfb^{l}$, where $\bfa^{l}$ and $\bfb^{l}$ are the outputs of $\mathcal{A}^{l}$ and $\mathcal{B}^{l}$ respectively. \cref{fig:logical_layer_example} illustrates an example of two adjacent logical layers, where the directed arrows indicates the presence of the connections between nodes. By examining the weights $W^{(l,0)}$ and $W^{(l,1)}$, we could derive the corresponding rules in both conjunctive and disjunctive normal forms.

\subsection{Training Paradigm}
While the logical layers are capable of expressing Boolean operations, their non-differentiable structure makes CRL challenging to optimize. To address this issue, we introduce continuous logical layers in training. These layers are differentiable and share the same parameters as the discrete counterparts, enabling end-to-end optimization while preserving the interpretability of the original design.

Let $\bar{W}^{(l,0)}, \bar{W}^{(l,1)}\in[0,1]$ denote the continuous weight matrices of conjunction and disjunction layers.  To make \cref{eq:discrete_logic} differentiable, we leverage the logical activation functions introduced by~\cite{laf}:
\begin{equation*}
\label{eq:LAF}
\textit{Conj}(\bfn,
\bar{W}_{i})= P(\prod_{j=1}^{N}F_{c}(n_{j}, \bar{W}_{i,j})), \;\; \textit{Disj}(\bfn, \bar{W}_{i})=1-P(\prod_{j=1}^{N}F_{d}(n_{j}, \bar{W}_{i,j})),\end{equation*}
where $N$ is the number of nodes, $F_{c}(n,w)=1-w(1-n)$ and $F_{d}(n,w)=1-n\cdot w$. If $\bfn$ and $W_{i}$ are both binary vectors, then $\textit{Conj}(\bfn, W_{i})=\bigwedge_{W_{i,j}=1}\bfn_{j}$ and $\textit{Disj}(\bfn, W_{i})=\bigvee_{W_{i,j}=1}\bfn_{j}$. $P(x)=1/(1-\log x)$ is a projection function to prevent gradients vanishing.
After using continuous weights and logical activation functions, the nodes in continuous logical layers are defined as follows:
\begin{equation*}
\label{eq:conj_hat_disj_hat}
\bar{a}_{i}^{l} =\textit{Conj}(\bar{\bfn}^{l-1}, \bar{W}_{i}^{(l,0)}), \;\;\;\;\;\; 
\bar{b}_{i}^{l}=\textit{Disj}(\bar{\bfn}^{l-1}, \bar{W}_{i}^{(l,1)}).
\end{equation*}

By employing continuous logical layers, CRL facilitates end-to-end training. The computational graph of CRL is illustrated in \cref{fig:compute_graph}. As shown, the concept predictor generates concept activations $\hat{\bfc} = g(\bfx)$ from input images. The binarized concept activations are passed to both the discrete and continuous logical layers to generate the task predictions given by $\hat{\bfy}=f(r(q(\hat{\bfc}))|W)$ and $\bar{\bfy}=f(r(q(\hat{\bfc}))|\bar{W})$. The objective function is defined as:
\begin{equation}
\label{eq:loss}
\calL = \calL_{t}(\hat{\bfy}, y) + \calL_{c}(\hat{\bfc}, \bfc) + \lambda||\bar{W}||_2,
\end{equation}
where $\calL_t$ is the cross-entropy loss, $\calL_c$ represents the mean cross-entropy loss across all training concepts and $\lambda$ is the regularization hyperparameter that controls the complexity of the logical layers. For backpropagation, as indicated by the dotted arrow in \cref{fig:compute_graph}, the gradients $\partial \calL_t / \partial \hat{\bfy}$ are grafted onto the backward pass of $f(r(\cdot|\bar{W}))$. To address the non-differentiability of the binarizer, we utilize a Straight-Through Estimator (STE)~\cite{ste}, which approximates the gradients for the discretization.

\subsection{Decision Interpretation}
As illustrated in the dashed box of \cref{fig:method_overview}, CRL provides concept explanations for individual predictions as well as global logical rules for the entire dataset, thereby integrating both local and global interpretability. After training, the weights of logical layers $r(\cdot)$ are analyzed to summarize $R$ logical rules, where $R$ depends on the number of nodes and the hyperparameter $\lambda$ in \cref{eq:loss}. During inference, the concept predictor $g(\cdot)$ first extracts concepts from images and matches them against the learned rules. Each rule is associated with a set of class-specific weights from the linear layer $f(\cdot)$. When a rule is triggered by a match (bold in \cref{fig:method_overview}), its corresponding weights are added to the overall class logits $\hat{\bfy}$. The final prediction is determined by the cumulative logits, which aggregate the contribution of all matched rules.

\section{Experiments}
\subsection{Experimental Setup}
\subsubsection{Tasks and datasets:}
To evaluate the proposed method across different medical scenarios, we assess two tasks: skin disease diagnosis and white blood cell (WBC) classification. \textbf{Skin disease diagnosis:} We employ the Fitzpatrick17k (\texttt{F17k}) dataset~\cite{fitzpatrick} and the Diverse Dermatology Images (\texttt{DDI}) dataset~\cite{ddi}, incorporating concept annotations from the SkinCon dataset~\cite{skincon}. The SkinCon dataset comprises 48 concepts annotated by board-certified dermatologists. Following the approach in~\cite{align_cbm}, we focus our training and testing to images labeled as \textit{Benign} (Ben) or \textit{Malignant} (Mal). \textbf{WBC classification:} We utilize the \texttt{PBC} dataset~\cite{pbc}, along with the concept annotations from the WBCAtt dataset~\cite{wbcatt}. The WBCAtt dataset contains 24 morphological attributes, and the classification includes five distinct classes.

\subsubsection{Implementation details:}
We employ ResNet-34 pretrained on the ImageNet dataset, as the backbone network. The balance parameter between concept and task loss is set to 1. We use the AdamW optimizer with an initial learning rate of $5\times10^{-5}$ and a weight decay of 0.01. The learning rate decays to zero following a cosine scheduler. Models are trained for 300 epochs with a batch size of 64. CRL is implemented with two logical layers, each comprising 256 nodes. The hyperparameter $\lambda$ set to $5\times10^{-6}$ to control the complexity of the rules.
For evaluation, we report the average accuracy (ACC) and F1 score (F1) for both concept prediction and diagnosis tasks. For skin disease diagnosis, we perform 5-fold cross-validation. For WBC classification, we adopt the original data split from~\cite{wbcatt}, conducting experiments with three different random seeds.

\subsection{Results}

\begin{table}[t]
\caption{Performance comparison on skin disease diagnosis and WBC classification tasks. \textbf{Bold} text indicates the best results, while \underline{underlined} text denotes the second-best results. The symbol $\star$ denotes methods employ binary concept values, while $\dagger$ indicates methods with global interpretability.}\label{tab:utility}
\centering
\resizebox{0.80\linewidth}{!}{
\begin{tabular}{l|l|C{2cm}C{2cm}|C{2cm}C{2cm}}
    \toprule
    \multirow{2}{*}{Dataset} &\multirow{2}{*}{Method} &\multicolumn{2}{c|}{Concept Metric} &\multicolumn{2}{c}{Diagnosis Metric} \\
     & &  ACC(\%) & F1(\%) &  ACC(\%) & F1(\%)\\
    \midrule
    \multirow{6}{*}{\texttt{F17k}} 
    & CBM~\cite{cbm} & 91.41$_{\pm 0.19}$ & \textbf{59.44}$_{\pm 0.97}$ & \underline{76.37}$_{\pm 3.45}$ & \underline{76.30}$_{\pm 3.45}$ \\  
    & align-CBM~\cite{align_cbm} & 89.77$_{\pm 0.52}$ & 58.98$_{\pm 1.41}$ & 75.93$_{\pm 2.39}$ & 75.83$_{\pm 2.41}$ \\  
    & CEM~\cite{cem} & 91.85$_{\pm 0.55}$ & \underline{59.11}$_{\pm 2.06}$ & 76.26$_{\pm 2.59}$ & 76.17$_{\pm 2.52}$ \\  
    & evi-CEM~\cite{evi_cem} & \underline{92.04}$_{\pm 0.80}$ & 58.65$_{\pm 1.71}$ & \textbf{76.47}$_{\pm 1.69}$ & \textbf{76.39}$_{\pm 1.66}$ \\  
    & hard-CBM$^{\star}$ & 86.36$_{\pm 1.09}$ & 49.57$_{\pm 0.48}$ & 73.51$_{\pm 2.94}$ & 73.43$_{\pm 2.92}$ \\
    & DCR$^{\dagger}$~\cite{dcr} & 91.03$_{\pm 0.80}$ & 49.40$_{\pm 1.00}$ & 75.05$_{\pm 2.12}$ & 74.97$_{\pm 2.12}$ \\
    & \cellcolor{lightgray}\textbf{CRL$^{\star\dagger}$} & \cellcolor{lightgray}\textbf{92.80}$_{\pm 0.47}$ & \cellcolor{lightgray}52.39$_{\pm 0.28}$ & \cellcolor{lightgray}75.95$_{\pm 3.09}$ & \cellcolor{lightgray}75.90$_{\pm 3.08}$ \\  
    \midrule
    \multirow{6}{*}{\texttt{PBC}} 
    & CBM~\cite{cbm} & \underline{95.21}$_{\pm 0.10}$ & \textbf{91.65}$_{\pm 0.15}$ & 98.93$_{\pm 0.14}$ & 98.44$_{\pm 0.22}$ \\
    & align-CBM~\cite{align_cbm} & 95.01$_{\pm 0.23}$ & 89.95$_{\pm 0.59}$ & 99.14$_{\pm 0.05}$ & \underline{99.25}$_{\pm 0.52}$ \\
    & CEM~\cite{evi_cem} & 94.98$_{\pm 0.34}$ & \underline{90.86}$_{\pm 0.69}$ & \underline{99.43}$_{\pm 0.12}$ & 99.23$_{\pm 0.18}$ \\
    & evi-CEM~\cite{evi_cem} & 94.44$_{\pm 0.91}$ & 89.18$_{\pm 2.09}$ & \textbf{99.57}$_{\pm 0.03}$ & \textbf{99.42}$_{\pm 0.03}$ \\
    & hard-CBM$^{\star}$ & 64.53$_{\pm 2.42}$ & 56.24$_{\pm 2.74}$ & 98.22$_{\pm 0.48}$ & 97.52$_{\pm 0.73}$ \\
    & DCR$^{\dagger}$~\cite{dcr} & 92.79$_{\pm 0.43}$ & 82.92$_{\pm 0.64}$ & 98.93$_{\pm 0.32}$ & 98.47$_{\pm 0.49}$ \\
    & \cellcolor{lightgray}\textbf{CRL$^{\star\dagger}$} & \cellcolor{lightgray}\textbf{95.32}$_{\pm 0.26}$ & \cellcolor{lightgray}90.51$_{\pm 0.73}$ & \cellcolor{lightgray}98.67$_{\pm 0.25}$ & \cellcolor{lightgray}98.03$_{\pm 0.42}$ \\
    \bottomrule
\end{tabular}
}
\end{table}

\subsubsection{Model utility analysis:}
To showcase the classification utility of the proposed method, we compare CRL with other concept-based methods on F17k and PBC datasets. The comparative methods include CBM~\cite{cbm}, align-CBM~\cite{align_cbm}, CEM~\cite{cem}, evi-CEM~\cite{evi_cem} and DCR~\cite{dcr}. Among these, align-CBM integrates clinical knowledge to prioritize the most relevant, while evi-CEM employs evidential learning to model concept uncertainty. Hard-CBM, which serves as a baseline, only accepts binary concepts without logical layers. The comparison results are presented in \cref{tab:utility}. From the results, we observe that CRL achieves comparable predictive performance with other CBM variants, though it exhibits a performance trade-off compared to methods based on concept embeddings. Notably, when comparing CRL and hard-CBM, both of which utilize binary concepts, CRL significantly outperforms hard-CBM. This improvement can be attributed to the logical layers in CRL, which capture the concept correlations with logical operations.

\subsubsection{Model generalizability analysis:}
\begin{table}[t]
\caption{Performance comparison on the unseen \texttt{DDI} dataset. \textbf{Bold} text indicates the best results, while \underline{underlined} text denotes the second-best results. The symbol $\star$ denotes methods employ binary concept values, while $\dagger$ indicates methods with global interpretability.}\label{tab:generalizability}
\centering
\resizebox{\linewidth}{!}{        
\begin{tabular}{l|C{1.8cm}C{1.8cm}C{1.8cm}C{1.8cm}|C{1.3cm}C{1.3cm}C{1.3cm}C{1.3cm}}
    \toprule
    \multirow{3}{*}{Method} &\multicolumn{4}{c|}{OOD Performance$\uparrow$} &\multicolumn{4}{c}{Performance Drop$\downarrow$} \\
    &\multicolumn{2}{c}{Concept Metric} &\multicolumn{2}{c|}{Diagnosis Metric} &\multicolumn{2}{c}{Concept Metric} &\multicolumn{2}{c}{Diagnosis Metric} \\
     &  ACC(\%) & F1(\%) &  ACC(\%) & F1(\%) &  ACC(\%) & F1(\%) &  ACC(\%) & F1(\%)\\
    \midrule
    CBM~\cite{cbm} & 90.59$_{\pm 0.27}$ & \underline{53.09}$_{\pm 0.44}$ & 62.36$_{\pm 4.05}$ & 57.49$_{\pm 2.02}$ & 0.82 & 6.35 & 14.01 & 18.81 \\
    align-CBM~\cite{align_cbm} & 89.34$_{\pm 0.25}$ & \underline{51.47}$_{\pm 0.22}$ & 63.13$_{\pm 4.01}$ & \underline{59.66}$_{\pm 2.18}$ & 0.33 & 7.51 & 12.80 & \underline{16.17} \\
    CEM~\cite{cem} & \underline{91.66}$_{\pm 0.26}$ & \textbf{53.21}$_{\pm 0.57}$ & 62.39$_{\pm 2.59}$ & 57.42$_{\pm 2.01}$ & \underline{0.19} & 5.90 & 13.87 & 18.75 \\
    evi-CEM~\cite{evi_cem} & 91.35$_{\pm 0.20}$ & 52.17$_{\pm 0.64}$ & \underline{64.25}$_{\pm 2.04}$ & 58.82$_{\pm 1.62}$ & 0.69 & 6.48 & 12.12 & 17.48 \\
    hard-CBM$^{\star}$ & 84.63$_{\pm 1.20}$ & 48.48$_{\pm 0.28}$ & 63.77$_{\pm 2.35}$ & 56.96$_{\pm 1.52}$ & 1.73 & \underline{1.09} & \underline{9.74} & 16.47 \\
    DCR$^{\dagger}$~\cite{dcr} & 90.97$_{\pm 0.49}$ & 47.89$_{\pm 0.72}$ & 62.23$_{\pm 4.09}$ & 57.50$_{\pm 2.35}$ & \textbf{0.06} & 1.51 & 12.82 & 17.47 \\
    \cellcolor{lightgray}\textbf{CRL$^{\star\dagger}$} & \cellcolor{lightgray}\textbf{92.14}$_{\pm 0.20}$ & \cellcolor{lightgray}51.78$_{\pm 0.19}$ & \cellcolor{lightgray}\textbf{73.46}$_{\pm 2.36}$ & \cellcolor{lightgray}\textbf{63.42}$_{\pm 1.27}$ & \cellcolor{lightgray}0.66 & \cellcolor{lightgray}\textbf{0.61} & \cellcolor{lightgray}\textbf{2.49} & \cellcolor{lightgray}\textbf{12.48} \\
    \bottomrule
\end{tabular}
}
\end{table}

To evaluate generalizability, we assess the out-of-domain (OOD) performance of the models on the unseen \texttt{DDI} dataset, using models trained on the source \texttt{F17k} dataset. As reported in \cref{tab:generalizability}, we can observe that concept models relying on soft concepts (probabilities or embeddings) exhibit relatively large performance drops, whereas methods employing binary concept values experience smaller declines. This suggests that binary concepts could help mitigate concept leakage. Notably, although both CRL and hard-CBM utilize binary concepts, CRL achieves a diagnostic ACC of 73.46, outperforming the second-best method by 9.21 while exhibiting the smallest performance drop. This improvement can be attributed to the domain-invariant logical rules extracted by logical layers, which enhance robustness to distribution shifts.

\subsubsection{Model interpretability analysis:}

\begin{figure}[t]
  \centering
   \includegraphics[width=1.0\linewidth]{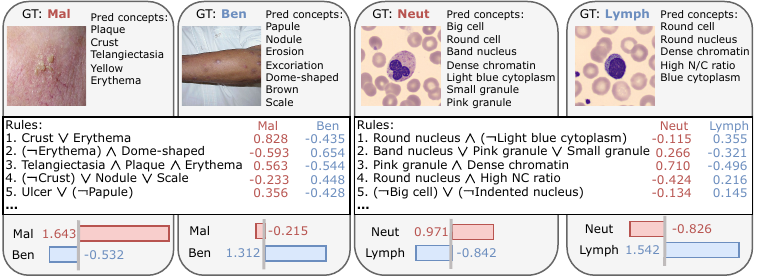}
   \caption{An illustration of concept logical rules for both skin disease diagnosis task (left) and WBC classification (right). Note: We only present the rule weights and logits of \textit{Neutrophil} (Neut) and \textit{Lymphocyte} (Lymph) for WBC classification.}
   \label{fig:rule_explanation}
\end{figure}

To illustrate that CRL could generate meaningful logical rules, we present the rules obtained from both skin disease diagnosis and WBC classification tasks, as illustrated in~\cref{fig:rule_explanation}. For skin disease diagnosis, we observe that concepts \texttt{Telangiectasia}, \texttt{Ulcer} and \texttt{Crust} are associated with \textit{Malignant}, aligning with established clinical knowledge~\cite{dermatology}. In case of WBC classification, the rules indicate that \textit{Lymphocytes} are characterized by a \texttt{High NC ratio} and a \texttt{Round nucleus}, while \textit{Neutrophils} typically exhibit \texttt{Pink granules} and \texttt{Dense chromatin}, consistent with clinical observations~\cite{wbcatt}. The case studies demonstrate that CRL can effectively extract concept-based logical rules that are clinically meaningful, offering both local concept explanations and global rule explanations for the entire medical dataset.

\section{Conclusion}
This paper introduces CRL, a framework for interpretable medical image classification that mitigates concept leakage and unifies local and global interpretability. By employing binary concepts and learnable logical layers, CRL effectively models concept correlations and extracts clinically meaningful decision rules. Experiments on two medical image classification tasks demonstrate that CRL achieves competitive performance and exhibits superior generalizability to unseen data.

\bibliographystyle{splncs04}
\bibliography{reference}

\end{document}